\documentclass{article}
\usepackage{spconf,amsmath,graphicx}
\usepackage{color}
\usepackage[utf8]{inputenc}

\usepackage{breakcites}

\usepackage{xcolor,colortbl}

\title{Segmental Audio Word2Vec: Representing Utterances as Sequences of Vectors with Applications in Spoken Term Detection}
\name{Yu-Hsuan Wang, Hung-yi Lee, Lin-shan Lee}
\address{College of Electrical Engineering and Computer Science\\
         National Taiwan University\\
         \footnotesize \tt \{r04922167, hungyilee\}@ntu.edu.tw, lslee@gate.sinica.edu.tw}
\begin{document}
%
\fontsize{9pt}{10.62pt}\selectfont

\maketitle
\begin{abstract}
While Word2Vec represents words (in text) as vectors carrying semantic information, audio Word2Vec was shown to be able to represent signal segments of spoken words as vectors carrying phonetic structure information. Audio Word2Vec can be trained in an unsupervised way from an unlabeled corpus, except the word boundaries are needed. In this paper, we extend audio Word2Vec from word-level to utterance-level by proposing a new segmental audio Word2Vec, in which unsupervised spoken word boundary segmentation and audio Word2Vec are jointly learned and mutually enhanced, so an utterance can be directly represented as a sequence of vectors carrying phonetic structure information. This is achieved by a segmental sequence-to-sequence autoencoder (SSAE), in which a segmentation gate trained with reinforcement learning is inserted in the encoder. Experiments on English, Czech, French and German show very good performance in both unsupervised spoken word segmentation and spoken term detection applications (significantly better than frame-based DTW).
\end{abstract}
\begin{keywords}
recurrent neural network, autoencoder, reinforcement learning, policy gradient
\end{keywords}
%
\section{Introduction}
\label{sec:intro}

In natural language processing, it is well known that Word2Vec transforming words (in text) into vectors of fixed dimensionality is very useful in various applications, because those vectors carry semantic information\cite{mikolov2013distributed}\cite{le2014distributed}. In speech signal processing, it has been shown that audio Word2Vec transforming spoken words into vectors of fixed dimensionality\cite{chung2016audio}\cite{hsu2017learning} is also useful for example in spoken term detection or data augmentation\cite{chen2015query}\cite{hsu2017unsupervised}, because those vectors carry phonetic structure for the spoken words. It has been shown that this audio Word2Vec can be trained in a completely unsupervised way from an unlabeled dataset, except the spoken word boundaries are needed. The need for spoken word boundaries is a major limitation for audio Word2Vec, because word boundaries are usually not available for given speech utterances or corpora\cite{chung2014unsupervised}\cite{zhang2009unsupervised}. 

\begin{sloppypar}
Although it is possible to use some automatic processes to estimate word boundaries followed by the audio Word2Vec\cite{kamper2017segmental}\cite{wang2017gate}\cite{rasanen2014basic}\cite{qiao2008unsupervised}\cite{lee2012nonparametric}, it is highly desired that the signal segmentation and audio Word2Vec may be integrated and jointly learned, because in that way they may enhance each other. This means the machine learns to segment the utterances into a sequence of spoken words, and transform these spoken words into a sequence of vectors at the same time. This is the segmental audio Word2Vec proposed here: representing each utterance as a sequence of fixed-dimensional vectors, each of which hopefully carries the phonetic structure information for a spoken word. This actually extends the audio Word2Vec from word-level up to utterance-level. Such segmental audio Word2Vec can have plenty of potential applications in the future, for example, speech information summarization, speech-to-speech translation or voice conversion\cite{kong2006improved}. Here we show the very attractive first application in spoken term detection.
\end{sloppypar}

The segmental audio Word2Vec proposed in this paper is based on a $\textit{segmental sequence-to-sequence autoencoder}$ (SSAE) for learning a segmentation gate and a sequence-to-sequence autoencoder jointly. The former determines the word boundaries in the utterance, and the latter represents each audio segment with an embedding vector. These two processes can be jointly learned from an unlabeled corpus in a completely unsupervised way. During training, the model learns to convert the utterances into sequences of embeddings, and then reconstructs the utterances with these sequences of embeddings. A guideline for the proper number of vectors (or words) within an utterance of a given length is needed, in order to prevent the machine from segmenting the utterances into more segments (or words) than needed. Since the number of embeddings is a discrete variable and not differentiable, the standard back-propagation is not applicable\cite{bengio2013estimating}\cite{chung2016hierarchical}. The policy gradient for reinforcement learning\cite{williams1992simple} is therefore used. How these generated word vector sequences carry the phonetic structure information of the original utterances was evaluated with the real application task of query-by-example spoken term detection on four languages: English (on TIMIT), Czech, French, German (on GlobalPhone corpora)\cite{schultz2002globalphone}.

\section{Proposed Approach}
\label{sec:approach}

\subsection{Segmental Sequence-to-Sequence Autoencoder (SSAE)}
\label{sec:segmental_rae}
\begin{figure}[t]
           \centering
           \includegraphics[width=\linewidth]{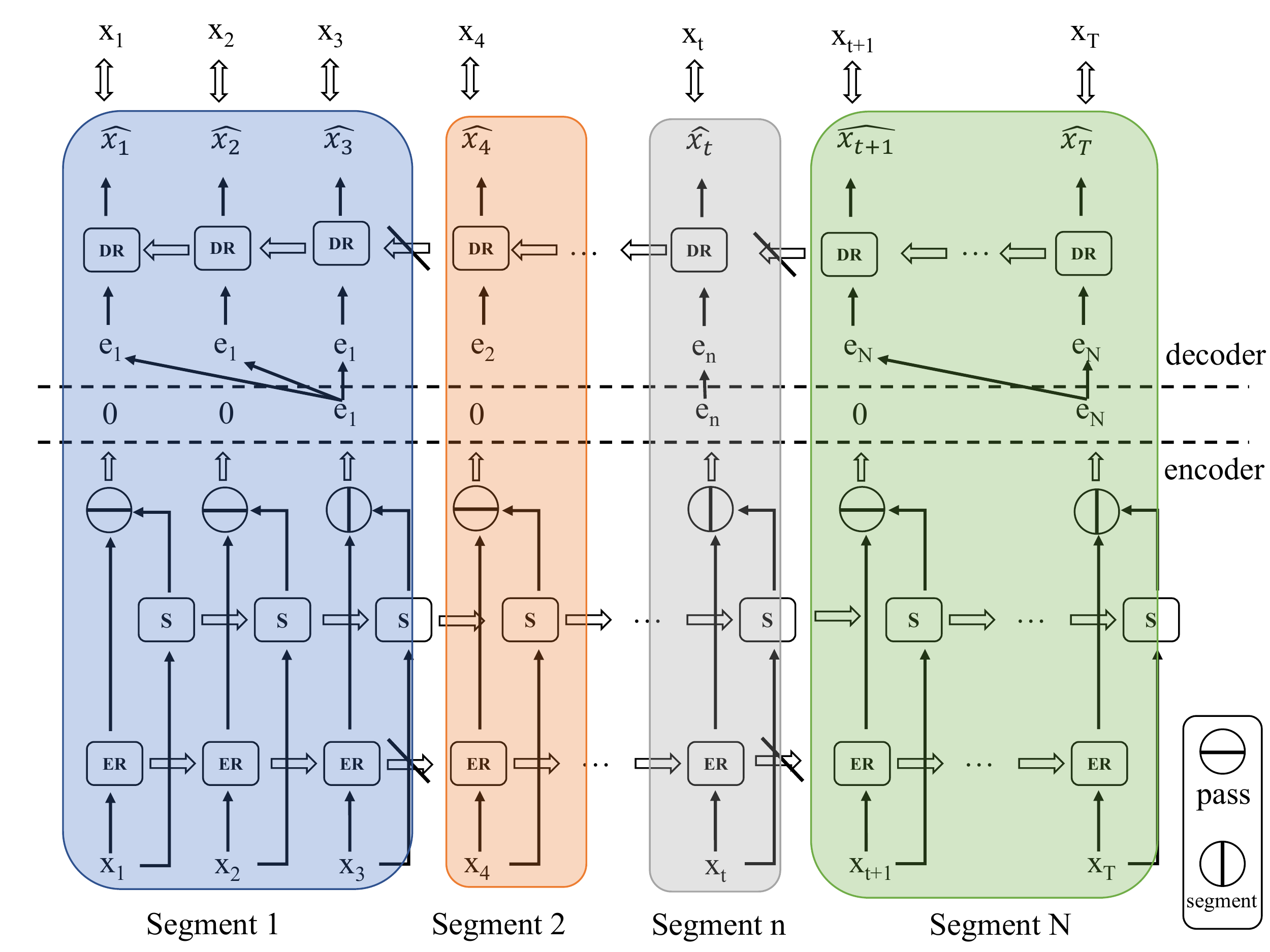}
           \caption{The segmental sequence-to-sequence autoencoder (SSAE). In addition to the encoder RNN (ER) and decoder RNN (DR), a segmentation gate (blocks $\textbf{S}$) is included in the encoder for estimating the word boundaries. During transitions across the segment boundaries, encoder RNN and decoder RNN are reset (illustrated with a slash in front of an arrow) so there is no information flow across segment boundaries. Each segment (shown in different colors) can be viewed as performing sequence-to-sequence training individually.}
           \label{fig:model_fig}
       \end{figure}

The proposed structure for SSAE is depicted in Fig.~\ref{fig:model_fig}, in which the $\textit{segmentation gate}$ is inserted into the recurrent autoencoder. For an input utterance $\textbf{X}$ $=$ \{$\mathbf{x}_1$, $\mathbf{x}_2$, ..., $\mathbf{x}_T$\}, where $\mathbf{x}_t$ represents the t-th acoustic feature like MFCC and $T$ is the length of the utterance, the model learns to determine the word boundaries and produce the embeddings for the $N$ generated audio segments, \textbf{Y} = \{ $\mathbf{e}_1, \mathbf{e}_2, ..., \mathbf{e}_N$\}, where $\mathbf{e}_n$ is the n-th embedding and $N \le T$.

The proposed SSAE consists of an encoder RNN (ER) and a decoder RNN (DR) just like the conventional autoencoder. But the encoder includes an extra segmentation gate, controlled by another RNN (shown as a sequence of blocks $\textbf{S}$ in Fig.~\ref{fig:model_fig}). The segmentation problem is formulated as a reinforcement learning problem. At each time $t$, the segmentation gate agent performs an action $a_t$, "segment" or "pass", according to a given state $\textbf{s}_t$. $\textbf{x}_t$ is taken as a word boundary if $a_t$ is "segment".

For the segmentation gate, the state at time $t$, $\textbf{s}_t$, is defined as the concatenation of the input $\mathbf{x}_t$, the gate activation signal (GAS) $\textbf{g}_t$ extracted from the gates of the GRU in another pre-trained RNN autoencoder \cite{wang2017gate}, and the previous action $a_{t-1}$ taken \cite{zoph2016neural},

        \begin{equation}
        \mathbf{s}_t = \big[ \mathbf{x}_t || \mathbf{g}_t || a_{t-1}\big].
        \label{eq:segment_gate_state}
        \end{equation}

The output $\mathbf{h}_t$ of layers of the segmentation gate RNN (blocks S in Fig.~\ref{fig:model_fig}) followed by a linear transform ($W^{\pi}$,$\mathbf{b}^{\pi}$) and a softmax nonlinearity models the policy $\pi_t$ at time t,

        \begin{equation}
        \mathbf{h}_t = \textit{RNN}(\textbf{s}_1,\textbf{s}_2,...\textbf{s}_t),
        \label{eq:srnn_rnn_output}
        \end{equation}
        \begin{equation}
        \pi_t = \textit{softmax}(W^{\pi}\mathbf{h}_t + \mathbf{b}^{\pi}).
        \label{eq:srnn_rnn_policy}
        \end{equation}
This $\pi_t$ gives two probabilities respectively for "segment" and "pass". An action $a_t$ is then sampled from this distribution during training to encourage exploration. During testing $a_t$ is "segment" whenever its probability is higher.

When $a_t$ is "segment", the time $t$ is viewed as a word boundary, and the segmentation gate passes the output of encoder RNN as an embedding. The state of the encoder RNN is also reset to its initial value. So the embedding $\mathbf{e}_n$ is generated based on the acoustic features of the audio segment only, independent of the previous input in spite of the recurrent structure,
         \begin{equation}
           \mathbf{e}_n = Encoder(\mathbf{x}_{t_1}, \mathbf{x}_{t_1 + 1}, ..., \mathbf{x}_{t_2}),
       \label{eq:embedding}
        \end{equation}       
     where $t_1$, $t_2$ refers to the beginning and ending time for the n-th audio segment.

The input utterance \textbf{X} should be reconstructed with the embedding sequence \textbf{Y} = \{ $\mathbf{e}_1, \mathbf{e}_2, ..., \mathbf{e}_N$\}. Because the decoder RNN (DR) is backward in order as shown in Fig.~\ref{fig:model_fig} \cite{sutskever2014sequence}, for the embedding $\mathbf{e}_n$ for the input segment from $t_1$ to $t_2$ in Eq.(\ref{eq:embedding}) above, the reconstructed feature vector is,
       \begin{equation}
          \mathbf{\hat{x}}_t = Decoder( \mathbf{\hat{x}}_{t_2}, \mathbf{\hat{x}}_{{t_2}-1}, ...\mathbf{\hat{x}}_{t+1},\mathbf{e}_n ).
        \label{eq:s-rae decoding}
        \end{equation} 

           The decoder RNN is also reset when beginning decoding each audio segment to remove the information flow from the following segment. 

\subsection{Encoder and Decoder Training}
\label{sec:encoder_and_decoder_training}
The loss function $\mathcal{L}$ for training the encoder and decoder is simply the averaged squared $\ell$-2 norm for the reconstruction error of all input $\mathbf{x}_t$:
       \begin{equation}
        \mathcal{L} = \sum_{l}^{L}\sum_{t}^{T_l}\frac{1}{d}\left \| \mathbf{\hat{x}}_{t}^{(l)} - \mathbf{x}_{t}^{(l)} \right \|^{2},
        \label{eq:srae_loss_eq}
        \end{equation}
        where the superscript $(l)$ indicates the $l$-th training utterance with length $T_l$, and $L$ is the number of utterances used in training. $d$ is the dimensionality of $\mathbf{x}_t^{(l)}$. 

\subsection{Segmentation Gate Training}
\label{sec:segmentation_gate}
\subsubsection{Reinforcement Learning}
The segmentation gate is trained with the reinforcement learning. After the segmentation gate performs the segmentation for each utterance, it receives a reward $r$ and a reward baseline $r_b$ for the utterance for updating the parameters. $r$ and $r_b$ will be defined in the next subsection.
We can write the expected reward for the gate under policy $\pi$ as $J(\theta) = \mathbf{E}_\pi[r]$, where $\theta$ is the parameter set. The updates of the segmentation gate are simply given by:
        \begin{equation}
        \nabla_\theta J(\theta) = \mathbf{E}_{a\sim{\pi}}[\nabla_\theta \sum_{t=1}^{T}log\pi^{(\theta)}_t(a_t)(r - r_b)],
        \label{eq:segmentor_rnn_update_rule}
        \end{equation}
where $\pi^{(\theta)}_t(a_t)$ is the probability for the action $a_t$ taken as in Eq.(\ref{eq:srnn_rnn_policy}).

\subsubsection{Rewards}
The reconstruction error is certainly a good indicator to see whether the segmentation boundaries are good, since the embeddings are generated based on the segmentation. We hypothesize that good boundaries, for example those close to word boundaries, would result in smaller reconstruction errors, because the audio segments for words would appear more frequently in the corpus and thus the embeddings would be trained better giving lower reconstruction errors. So the smaller the reconstruction errors the higher the reward:
        \begin{equation}
        r_{\textit{MSE}} = -\sum_{t}^{T}\frac{1}{d}\left \| \mathbf{\hat{x}}_{t} - \mathbf{x}_{t} \right \|^{2}.
        \label{eq:re_loss_reward}
        \end{equation}
This is very similar to Eq.(\ref{eq:srae_loss_eq}) except for a specific utterance here.

On the other hand, a guideline for the proper number of segments (words) $N$ in an utterance of a given length $T$ is important, otherwise for minimizing the reconstruction error as many segments as possible will be generated. So the smaller number of segments $N$ normalized by the utterance length $T$, the higher the reward:

        \begin{equation}
        r_{\textit{N/T}} = -\frac{N}{T},
        \label{eq:embed_num_reward}
        \end{equation}      
where $N$ and $T$ are respectively the numbers of segments and frames for the utterance as in Fig.~\ref{fig:model_fig}.

The total rewards $r$ is obtained by choosing the minimum between $r_{\textit{MSE}}$ and $r_{\textit{N/T}}$:
        \begin{equation}
        r = min(r_{\textit{MSE}}, \lambda r_{N/T})
        \label{eq:segmentor_rnn_reward}
        \end{equation}
where $\lambda$ is a hyperparameter to be tuned for a reasonable guideline for estimating the proper number of segments for an utterance of length $T$. In our experiments, this minimum function gave better results than linear interpolation.

We further use utterance-wise reward baseline to remove the bias between utterances. For each utterance, $M$ different sets of segment boundaries are sampled by the segmentation gate, each used to evaluate a reward $r_m$ with Eq.(\ref{eq:segmentor_rnn_reward}). The reward baseline $r_b$ for the utterance is then the average of them:

        \begin{equation}
        r_b = \frac{1}{M}\sum_{m=1}^{M}r_m.
        \label{eq:segmentor_rnn_reward_baseline}
        \end{equation}   

\subsection{Iterative Training Process}
Although all the models described in sections \ref{sec:encoder_and_decoder_training} and \ref{sec:segmentation_gate} can be trained simultaneously, we actually trained our model with an iterative process consisting of two phases. 
The first phase is to train the encoder and decoder with Eq.(\ref{eq:srae_loss_eq}) while fixing the parameters of the segmentation gate. The second phase is to update the parameters of the segmentation gate with rewards provided by the encoder and decoder while fixing their parameters. The two phases are performed iteratively. In phase one, the encoder and decoder should be learned from random initialized parameters each time, instead of taking the parameters learned in the previous iteration as the initialization, which was found to offer better training stability.

\section{Example Application: Unsupervised Query-by-example Spoken Term Detection}
\label{sec:approach_std}

\begin{figure}[t]
           \centering
           \includegraphics[width=\linewidth]{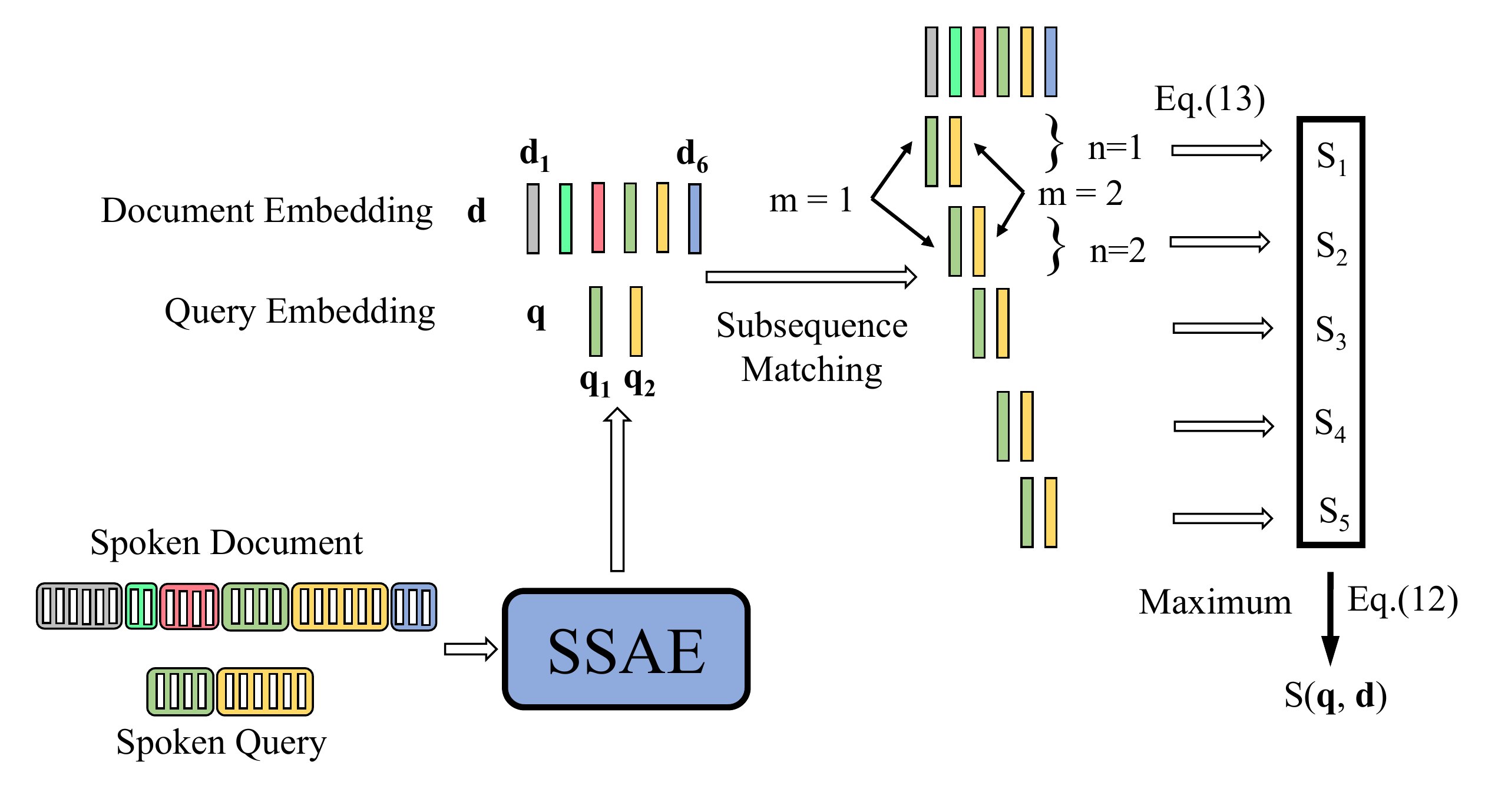}
           \caption{An illustration of unsupervised spoken term detection performed with segment-based subsequence matching using SSAE.}
           \label{fig:std_approach}
       \end{figure}

This approach can be used in many potential applications. Here we consider the unsupervised query-by-example spoken term detection (QbE STD) as the first example application. The task of unsupervised QbE STD is to locate the occurrence regions of the input spoken query in a large spoken archive without performing speech recognition. With the SSAE proposed here, this can be achieved as illustrated in  Fig.~\ref{fig:std_approach}. Given frame sequences of a spoken query and a spoken document, SSAE can represent these sequences as embeddings, $\textbf{q}$ = \{ $\mathbf{q}_1$, $\mathbf{q}_2$, ..., $\mathbf{q}_{N_q}$\} for the query and $\textbf{d}$ = \{ $\mathbf{d}_1$, $\mathbf{d}_2$, ..., $\mathbf{d}_{N_d}$  \} for the document. With the embeddings, simply subsequence matching can be used to evaluate the relevance score $S(\textbf{q},\textbf{d})$ between $\textbf{q}$ and $\textbf{d}$:
        \begin{equation}
         S(\mathbf{q}, \mathbf{d}) = max(S_1, S_2, ..., S_n, ..., S_{N_d - N_q + 1}),
        \label{eq:subsequent_matching_eq1}
        \end{equation}
        \begin{equation}
         S_n = \prod_{m=1}^{N_q}Sim(\mathbf{q}_m, \mathbf{d}_{m+n-1}).
        \label{eq:subsequent_matching_eq2}
        \end{equation}
Cosine similarity can be used in the similarity measure in Eq.(\ref{eq:subsequent_matching_eq2}). As is clear in the right part of Fig.~\ref{fig:std_approach}, $S_1$ = $sim(\mathbf{q}_1, \mathbf{d}_1)\cdot sim(\mathbf{q}_2, \mathbf{d}_2)$, $S_2$ = $sim(\mathbf{q}_1, \mathbf{d}_2)\cdot sim(\mathbf{q}_2, \mathbf{d}_3)$ and so on. The relevance score $S(\mathbf{q}, \mathbf{d})$ in Eq.(\ref{eq:subsequent_matching_eq1}) between the query and document is then the maximum out of all $S_n$'s obtained in Eq.(\ref{eq:subsequent_matching_eq2}). In this way, the frame-based template matching such as DTW can be replaced by segment-based subsequence matching with much less on-line computation requirements.        

\section{Experiments}
\label{sec:experiments}

\subsection{Experimental Setup}
\label{sec:experimental_setup}

We performed the experiments on four different languages: English, Czech, French, German. The English corpus was TIMIT and the corpus for the other languages was the GlobalPhone\cite{schultz2002globalphone}. The ground truth word boundaries for English were provided by TIMIT, while for the other three languages we used the forced aligned word boundaries. Both the encoder and decoder RNNs of the SSAE consisted of one hidden layer of 100 LSTM units\cite{hochreiter1997long}. The segmentation gate consisted of 2 layers of LSTM of size 256. All parameters were trained with Adam\cite{kingma2014adam}. $M=5$ in Eq.(\ref{eq:segmentor_rnn_reward_baseline}) in estimating the reward baseline for each utterance. The proximal policy optimization algorithm was used to train the reinforcement learning model\cite{schulman2017proximal}. The tolerance window for word segmentation evaluation was taken as 40 ms. The acoustic features used were 39-dim MFCCs with utterance-wise cepstral mean and variance normalization (CMVN) applied. In our experiments $\lambda = 5$ in Eq.(\ref{eq:segmentor_rnn_reward}), which was obtained empirically and obviously had to do with the average duration of the segmented spoken words. In spoken term detection, 5 words for each language containing a variety of phonemes were randomly selected to be the query words as listed in Table~\ref{table:query_word}, and several occurrences for each of them in training set were used as the spoken queries. The testing set utterances were used as spoken documents \cite{zhang2009unsupervised}. The numbers of spoken queries used for evaluation on English, Czech, French and German were 29, 21, 25 and 23 respectively.

\begin{table}[h]
\centering
\begin{tabular}{|l|c|}
\hline
 Language& Query Words  \\ \hline
 English & fail, simple, military, increases, problems \\ \hline  
 Czech& pracují, použití, textu, demokracie, abych \\ \hline
 French & soldats, organisme, travaillant, soulève, sportifs \\ \hline
 German& vergeblich, gutem, sozial, großes, ernennung \\ \hline
\end{tabular}
\caption{List of the randomly selected query words containing a variety of phonemes for English, Czech, French and German used in the experiments.}
\label{table:query_word}
\end{table}

\subsection{Spoken Word Segmentation Evaluation}
\label{sec:stage1_exp}
\begin{figure}[htb]  
\begin{minipage}[b]{.48\linewidth}
  \centering
  \centerline{\includegraphics[width=4.0cm]{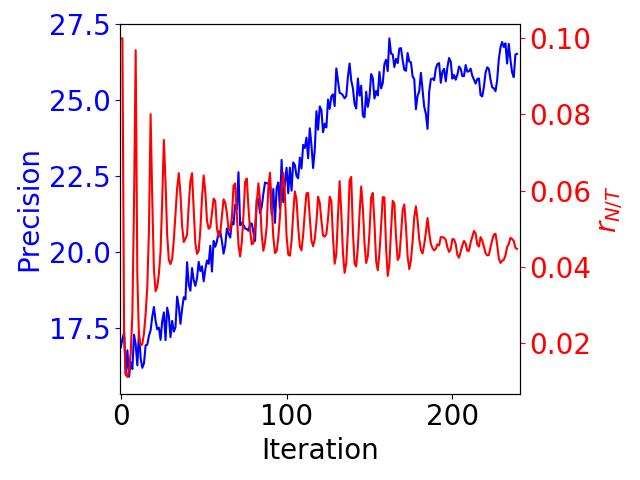}}
  \centerline{(a) Precision}\medskip
\end{minipage}
\hfill
\begin{minipage}[b]{0.48\linewidth}
  \centering
  \centerline{\includegraphics[width=4.0cm]{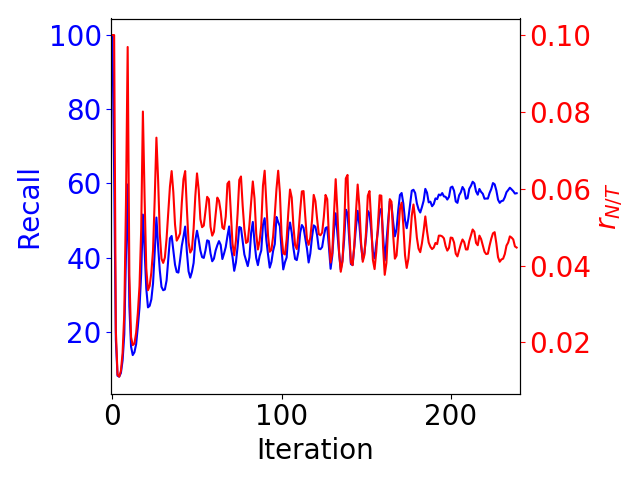}}
  \centerline{(b) Recall}\medskip
\end{minipage}
\caption{The learning curves of SSAE on the Czech validation set. The red curves are for $r_{N/T}$ in Eq.(\ref{eq:embed_num_reward}). The blue curves are (a) precision and (b) recall.}
\label{fig:stage1_czech_learning_curves}
\end{figure}

Fig.~\ref{fig:stage1_czech_learning_curves} shows the learning curves for SSAE on the Czech validation set. From the figure, we can see that SSAE gradually learned to segment utterances into spoken words because both the precision and recall (blue curves in Fig.~\ref{fig:stage1_czech_learning_curves}(a)(b) respectively) got higher when the reward $r_{N/T}$ in Eq.(\ref{eq:embed_num_reward}) (red curves) converged to a reasonable number. Similar trends were found in the other three languages. 

We evaluated the spoken word segmentation performance of the proposed SSAE by comparison with the random segmentation baseline and two segmentation methods, one using Gate Activation Signals (GAS)\cite{wang2017gate} and the other using the hierarchical agglomerative clustering (HAC)\cite{qiao2008unsupervised}\cite{chan2012unsupervised}, and the results are shown in Table~\ref{table:stage1_exp} in terms of F1 score. Precision (P) and recall (R) were also provided for English.
We see that the proposed SSAE performed significantly better than the other two methods on all languages except comparable to GAS for German. Also, for English recall about 50\% was achieved while the precision was significantly lower, which implies many of the word boundaries were actually identified, but many spoken words were in fact segmented into subword units. Similar trends were found for other languages.

    \begin{table}[th]   
      \centering
      \begin{tabular}{|l|c|c|c|c|c|c|}
        \hline
           {Lang.}&\multicolumn{3}{|c|}{English}& CZ & FR & GE\\ \hline
           Method&P&R&F1&\multicolumn{3}{|c|}{F1}\\ \hline
           Random&24.60&41.08&30.77&22.56&32.66&25.41\\ \hline
           HAC & 26.84 &46.21 & 33.96&30.84&33.75&27.09\\ \hline
           GAS & 33.22 &52.39&40.66 & 29.53&31.11&32.89\\ \hline
           \cellcolor{blue!25}SSAE&\cellcolor{blue!25}37.06&\cellcolor{blue!25}51.55&\cellcolor{blue!25}43.12&\cellcolor{blue!25}37.78&\cellcolor{blue!25}48.14&\cellcolor{blue!25}31.69\\ \hline       
      \end{tabular}
     \caption{Spoken word segmentation evaluation compared to different methods across different languages. Random segmentation was also provided as a baseline.}
     \label{table:stage1_exp}
    \end{table}  

\subsection{Spoken Term Detection (STD) Evaluation}
\label{sec:stage2_exp}
We evaluated the quality of embeddings generated by SSAE with the real application of spoken term detection using the method presented in section \ref{sec:approach_std}, compared with other kinds of audio Word2Vec embeddings trained with signal segments generated from different segmentation methods. Mean Average Precision (MAP) was used as the performance measure. 

The results are listed in Table~\ref{table:stage2_exp}. The performance for embeddings trained with ground truth word boundaries (oracle) in the last column serves as the upper bound. The random baseline in the first column simply assigned a random score to each pair of query and document. We also list the performance of standard frame-based dynamic time warping (DTW) as a primary baseline in the second column\cite{zhang2009unsupervised}. From the table, it is clear that the oracle achieved the best and significantly better performance than all other methods on all languages. SSAE outperformed the DTW baseline by a wide gap. This is probably because DTW may not be able to identify the spoken words if the speaker or gender characteristics are very different, but such different signal characteristics may be better absorbed in the audio Word2Vec training. These experimental results verified that the embeddings obtained with SSAE did carry the sequential phonetic structure information in the utterances, leading to the better performance in STD here. The performance of embeddings trained with GAS and HAC are not too far from random in most cases. It seems the performance of the spoken word segmentation has to be above some minimum level, otherwise the audio Word2Vec couldn't be reasonably trained, or spoken word segmentation boundaries had the major impact on the STD performance.

However, interestingly, although the segmentation performance of GAS was slightly better than SSAE for German, SSAE outperformed GAS a lot on spoken term detection for German. The reason is not clear yet, probably due to some special characteristics of the German language.

\begin{table}[th]
      \centering
      \begin{tabular}{|l|c|c|c|c|c|c|}
        \hline
        &&&\multicolumn{4}{|c|}{Embeddings (Different Seg.)} \\ \hline
        {Lang.}&Ran. &DTW& GAS& HAC&\cellcolor{blue!25}SSAE& Oracle\\ \hline
        Czech     &0.38& 16.59 &0.68&1.13&\cellcolor{blue!25}19.41&22.56 \\ \hline
        English   &0.74&12.02&8.29&0.91&\cellcolor{blue!25}23.27&30.28 \\ \hline
        French    &0.27& 11.72 &0.40&0.92&\cellcolor{blue!25}21.70& 29.66 \\ \hline
        German    &0.18&6.07&0.27&0.26&\cellcolor{blue!25}13.82&21.52\\ \hline
        \end{tabular}
       \caption{The spoken term detection performance in Mean Average Precision (MAP) for the proposed SSAE compared to the audio Word2Vec embeddings trained with spoken words segmented with other methods for different languages. The random baseline (Ran.) simply assigned a random score to each pair of query and document. Standard frame-based DTW is the primary baseline, while the oracle segmentation is the upper bound.}
       \label{table:stage2_exp}
    \end{table}

\section{Conclusion}
\label{sec:conclusion}
We propose in this paper the segmental sequence-to-sequence autoencoder (SSAE), which jointly learns and performs the spoken word segmentation and audio word embedding together. This actually extends the audio Word2Vec from word-level to utterance-level. This is achieved by reinforcement learning considering both the reconstruction errors obtained with the embeddings and the reasonable number of words within the utterances. Due to the reset mechanism in SSAE, an embedding is generated only based on an audio segment, therefore can be regarded as the audio word vector representing the segment. This is verified by the improved performance in experiments on unsupervised word segmentation and spoken term detection on four languages.

\vfill\pagebreak

\bibliographystyle{IEEEbib}
\bibliography{refs}

\end{document}